\documentclass[conference,compsoc]{IEEEtran}
\usepackage{amsmath,graphicx}
\usepackage{multirow}
\usepackage{placeins}

\usepackage[font=small,skip=0pt]{caption}
\usepackage{url}

\ifCLASSOPTIONcompsoc
  \usepackage[nocompress]{cite}
\else
  \usepackage{cite}
\fi

\ifCLASSINFOpdf
\else
\fi

\begin{document}
	\nocite{*}

\title{Slim LSTM NETWORKS: LSTM\_6 and LSTM\_C6}


\author{\IEEEauthorblockN{Atra Akandeh and Fathi M. Salem}
\IEEEauthorblockA{Circuits, Systems, and Neural Networks (CSANN) Laboratory \\
Computer Science and Engineering $||$ Electrical and Computer Engineering \\
University Neuroscience Program\\
Michigan State University\\
East Lansing, Michigan 48864-1226\\
akandeha@msu.edu; salemf@msu.edu }
}

\maketitle

\begin{abstract}
We have shown previously that our parameter-reduced variants of Long Short-Term Memory (LSTM) Recurrent Neural Networks (RNN) are comparable in performance to the \textit{standard LSTM RNN} on the MNIST dataset. In this study, we show that this is also the case for two diverse benchmark datasets, namely, the review sentiment IMDB and the 20 Newsgroup datasets. Specifically, we focus on two of the simplest variants, namely LSTM\_6 (i.e., \textit{standard} LSTM with three constant fixed gates) and LSTM\_C6 (i.e., LSTM\_6 with further reduced cell body input block). We demonstrate that these two aggressively reduced-parameter variants are competitive with the \textit{standard} LSTM when hyper-parameters, e.g., learning parameter, number of hidden units and gate constants are set properly. These architectures enable speeding up training computations and hence, these networks would be more suitable for online training and inference onto portable devices with relatively limited computational resources.
\end{abstract}
 

%
\IEEEpeerreviewmaketitle

\section{Introduction}
Recurrent Neural  Networks (RNNs) have been making an impact in sequence-to-sequence mappings, with particularly successful applications in speech recognition, music, language translation, and natural language processing to name a few 
\cite{Odyssey2016, zaremba2015empirical, chung2014empirical, boulanger2012modeling, DBLP:journals/corr/JohnsonSLKWCTVW16}.  By their structure, they possess a memory (or state) and include feedback or recurrence. The \textit{simple} RNN (sRNN) is succinctly expressed, see e.g., \cite{Goodfellow-et-al-2016}:

\begin{equation}
	\begin{split}
		& h_t = \sigma(W_{hx} x_t + W_{hh} h_{t-1} + b_h) \\
		& y_t = W_{hy} h_t + b_y
	\end{split}
\end{equation}
where $x_t$ is the input sequence vector at (time) step $t$, $h_t$ is the hidden (activation) unit vector at step $t$, while $h_{t-1}$ is the hidden unit vector at the previous step  $t-1$, and $y_t$ is the output vector at step $t$. The parameters are the three matrices, namely, $W_{hx}, W_{hh}$, and $W_{hy}$, and the vector $b_h$. This constitutes a discrete-step dynamic recurrent system with $h_t$ acting as the state. The parameters are to be determined adaptively via  training mostly using various versions of backpropagation through time (BPTT), e.g., see \cite{Odyssey2016}. 

The LSTM RNNs introduce a cell-memory and 3 gating signals to enable effective learning via the BPTT \cite{Odyssey2016}. The simple  activation state has been replaced with a more involved activation with gating mechanisms. The LSTM RNN uses a (additional) memory cell (vector and includes three gates: (i) an input gate, $i_t$  (ii) an output  gate $o_t$, and (iii) a forget gate, $f_t$. These gates collectively control signaling. The \textit{standard} LSTM is expressed mathematically as \cite{Odyssey2016, Goodfellow-et-al-2016}:

\begin{equation}
	\begin{split}
		& i_t = \sigma_{in}(W_i x_t + U_i h_{t-1} + b_i) \\
		& f_t = \sigma_{in}(W_f x_t + U_f h_{t-1} + b_f) \\
		& o_t = \sigma_{in}(W_o x_t + U_o h_{t-1} + b_o) \\
		& \tilde{c_t} = \sigma(W_c x_t + U_c h_{t-1} + b_c) \\
		& c_t= f_t \odot c_{t-1} + i_t \odot \tilde{c_t}\\
		& h_t = o_t \odot \sigma(c_t) 
	\end{split}
\end{equation}
where the first 4 equations are replica of the simple RNN (sRNN) above, with the first 3 equations serving as gating signals and thus their nonlinear activation is set as a sigmoid function $\sigma_{in}$, while the 4th equation's nonlinearity is an arbitrary nonlinearity $\sigma$, typically sigmoid, hyperbolic tangent ($tanh$), or rectified linear unit (reLU). This 4th equation is sometimes referred to as the input block. The last two equations entail the memory cell $c_t$ and now activation hidden unit $h_t$ with the insertion of the gating signals ina point-wise (Haramard) multiplications (using the symbol $\odot$). This represents a discrete-step nonlinear dynamic system with recurrence. The distinct parameters are associated with each replica as $W_*, U_*$, and $b_*$ is a straight fashion. 

The output layer of the LSTM model may be chosen to be as a linear (more accurately, affine) map as

\begin{equation}
	y_t = W_{hy} h_t + b_y
\end{equation}
where $y_t$ is the output, and $ W_{hy}$ is a matrix, and $b_y$ is a bias vector. In other optional implementation, this layer may be followed by a softmax layer to render the output analogous with probability ranges. 

LSTMs are relatively compationally expensive due to the fact that they have four replica with distinct sets of parameters (namely, weights and biases)  which would need to be adaptively updated every (mini-batch) of training calculations. 

We have introduced numerous, computationally simpler, LSTM variants by aggressively eleminating some of the adaptive parameters, see \cite{salem2016reduced, LuSalem2017, AkandehSalem2017I, AkandehSalem2017II, AkandehSalem2017III, slim-lstm, kent}. In this study we shall focus on one of the simplest variant forms, namely the slim LSTM\_6  and LSTM\_C6 
\cite{slim-lstm, AkandehSalem2017II}.

\section{LSTM\_6}
Different variants have been introduced earlier \cite{AkandehSalem2017III, slim-lstm}. For LSTM\_6, the gating signals are set at constant values as follows:

\begin{equation}
	\begin{split}
		& i_t = 1.0 \\
		& f_t = f,  ~-1 < f < 1.0 \\
		& o_t = 1.0 \\
		& \tilde{c_t} = \sigma(W_c x_t + U_c h_{t-1} + b_c) \\
		& c_t= f_t \odot c_{t-1} + i_t \odot \tilde{c_t}\\
		& h_t = o_t \odot \sigma(c_t) 
	\end{split}
\end{equation}

Note that the gate signal values are set to the constant scalars $f$ or $1$. In practice, when the gate is set to $1$, it is equivalent to eliminating the gate entirely!  Thus, in compact form, the {LSTM\_6} equation now reads: 
\begin{equation}
	\begin{split}
		& c_t= f~ c_{t-1} + \sigma(W_c x_t + U_c h_{t-1} + b_c) , ~~ -1 < f < 1.0 \\
		& h_t = \sigma(c_t) 
	\end{split}
\end{equation}
This variant form is close to the so-called basic Recurrent Neural Network (bRNN), see \cite{salem2016basic, slim-lstm} for analysis and details.

\section{LSTM\_C6}
In LSTM\_C6 the matrix $U_c$ in the cell equation is replaced with a corresponding vector $u_c$, in order to render a point-wise multiplication instead. This the variant equations become
\begin{equation}
	\begin{split}
		& i_t = 1.0 \\
		& f_t =f, ~ -1 < f < 1.0 \\
		& o_t = 1.0 \\
		& \tilde{c_t} = \sigma(W_c x_t + u_c  \odot h_{t-1} + b_c) \\
		& c_t= f_t \odot c_{t-1} + i_t \odot \tilde{c_t}\\
		& h_t = o_t \odot \sigma(c_t) 
	\end{split}
\end{equation}
Similarly, in compact form, these equations now read as: 
\begin{equation}
	\begin{split}
		& c_t= f ~ c_{t-1} +\sigma(W_c x_t + u_c  \odot h_{t-1} + b_c) , ~~ -1 < f < 1.0 \\
		& h_t = \sigma(c_t) 
	\end{split}
\end{equation}

To account for the number of parameters in each case, let the input vector $x_t$ be of $m$ dimension, the state $c_t$ and its activation hidden unit has dimension of $n$. Then the number of (adaptive) parameters in LSTM\_6 is $n(m+n+1)$ and for LSTM\_C6 the total number of (adaptive) parameters is $n(m+2)$. (Note that one may add to each the new nonadaptive \textit{hyper-parameter} $f$). Thus if the state dimension $n=100$ and the input dimension is $m=32$, the total number of (adaptive) parameters for LSTM\_6 is $3400$.

\FloatBarrier

Table \ref{vs-im} and Table \ref{vs-20} provide a summary of the number of parameters as well as the times per epoch during training corresponding to each of the model variants for $h=100$ for each data set. The number of parameter only include parameter corresponding  to LSTM layer and parameter of embedding and last dense layer is not included. These simulation and the training times are obtained by running the Keras Library \cite{keras_rowwise} with GPU option enable. Although, we expect that LSTM\_C6 takes less time per epoch than LSTM6, but due to Keras internal implementation, that is not the case. However, LSTM\_C6 is still faster than basic LSTM. Comparing these two table indicate that time-wise, parameter reduction plays a huge role in larger networks.

\begin{table}
	\caption{Variants specifications:  IMDB Dataset}
	\centering
	\begin{tabular}{| c | c |c|} 
		\hline
		variants & \# of parameters & dimensions \\ 
		\hline
		LSTM & 53200 & {m=32, n=100} \\ 
		\hline
		LSTM6 & 13300 & {m=32, n=100} \\ 
		\hline
		LSTM\_C6 & 3400 & {m=32, n=100} \\
		\hline
	\end{tabular}
	\label{vs-im}
\end{table}

\begin{table}
	\caption{Variants specifications: 20 Newsgroup Dataset}
	\centering
	\begin{tabular}{| c | c |c|} 
		\hline
		variants & \# of parameters & dimensions \\ 
		\hline
		LSTM & 263168& {m=128, n=128}  \\
		\hline
		LSTM6 & 65792 & {m=128, n=128} \\
		\hline
		LSTM\_C6 & 33280& {m=128, n=128}   \\
		\hline
	\end{tabular}
	\label{vs-20}
\end{table}

\section{Experiments and Discussion}
In the previous work \cite{AkandehSalem2017II}, we have shown that our networks are competitively comparable to \textit{standard} LSTM networks on the MNIST dataset. Here we show that LSTM\_6 (also denoted here as LSTM6) and LSTM\_C6 can compete with the \textit{standard} LSTM network in the benchmark public datasets IMDB and 20 Newsgroup available via the Keras library \textbf{https://keras.io}.

\subsection{The IMDB dataset}
IMDB Datasets is a binary sentiment classification dataset. To train the model, dictionary size of 5000 has been used. Each review is truncated or padded to 500 words. The first layer is an embedding layer which is a simple multiplication that transforms words into their corresponding word embedding. The output is then passed to an LSTM layer following a dense layer. The network specification which has been adopted from Keras 1.2 examples is given in table \ref{sp-im}.  

\begin{table}[ht]
	\caption{Network specifications: IMDB Datasets}
	\centering
	\begin{tabular}{| c | c |} 
		\hline
		Input dimension & $500 \times 32$ \\ 
		\hline
		Number of hidden units & $100,200,300,400$ \\
		\hline
		Non-linear function & sigmoid \\
		\hline
		Output dimension & 1 \\
		\hline
		Number of epochs & $ 100 $ \\
		\hline
		Optimizer  & Adam  \\
		\hline
		Batch size & $32$ \\
		\hline
		Loss function & binary cross-entropy\\
		\hline
	\end{tabular}
	\label{sp-im}
\end{table}

\begin{figure}[!htb]
	\centering
	\includegraphics[trim={0 0 0 0},clip,scale=0.48]{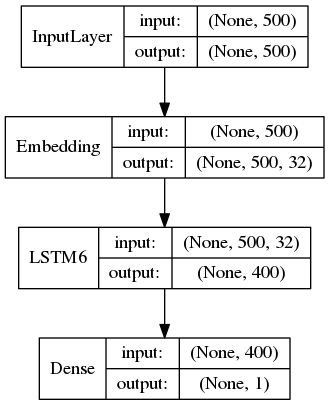}
	\vspace*{+2mm}
	\caption{Network Architecture Diagram for IMDB}
	\label{im-mdl}
\end{figure}

A schematic representation of the architecture used is given in figure \ref{im-mdl}.\\

In this experiment, the $sigmoid$ nonlinearity is used, since the $tanh$ nonlinearity has caused large fluctuations in training and testing outcomes and using the $reLU$ nonlinearity routinely failed to converge even for the \textit{standard} LSTM RNN. 

\subsubsection{Tuning the hyper-parameter $\eta$}
\begin{figure}[!htb]
	\centering
	\includegraphics[trim={0 0 0 0},clip,scale=0.42]{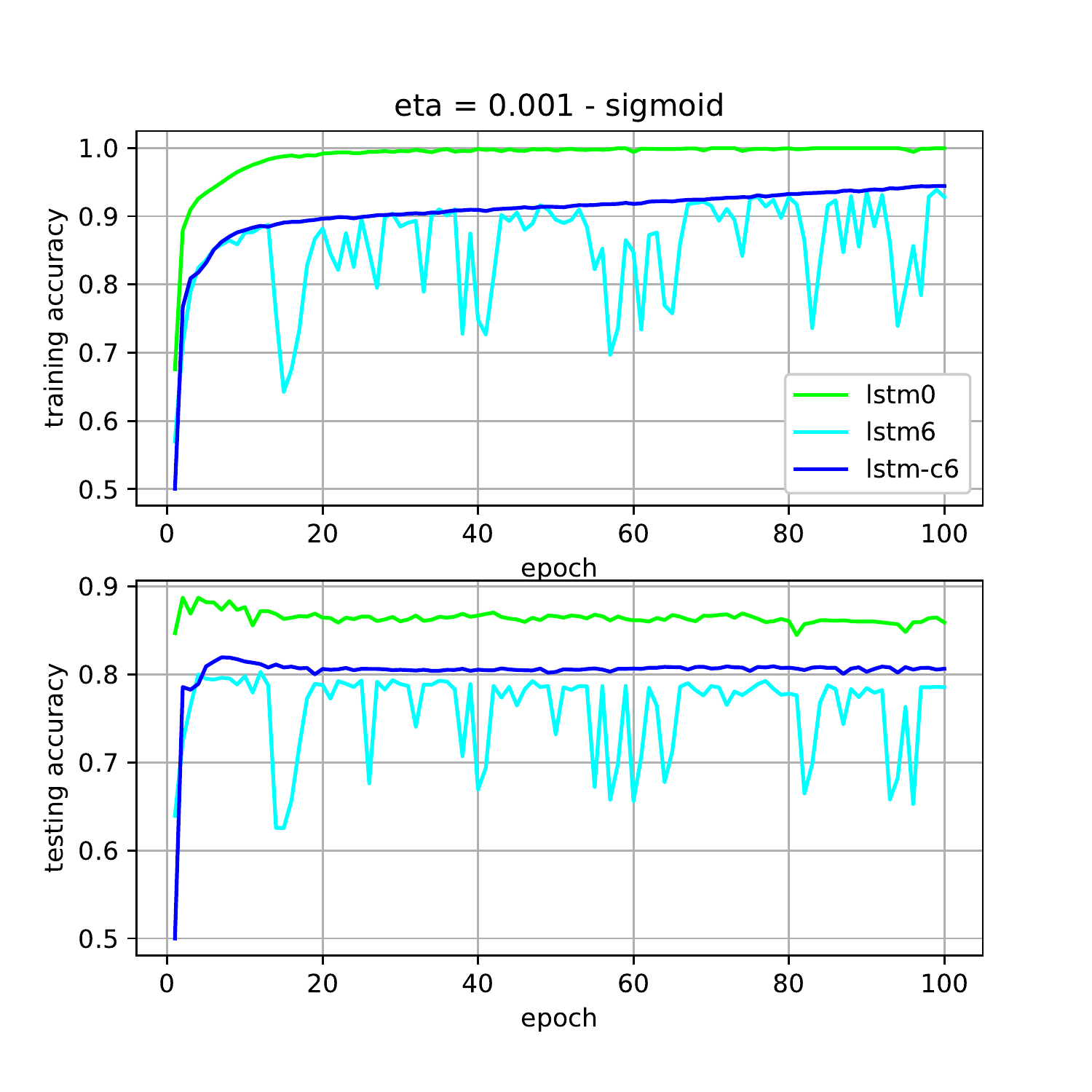}
	\caption{IMDB, Training \& Test accuracy, $\sigma=sigmoid , \eta=1\mathrm{e}{-3}$}
	\label{im001}
\end{figure}

We started with the generic $\eta = 1\mathrm{e}{-3}$ as used in the Keras library example. As it is shown in figure \ref{im001}, the \textit{standard} LSTM (denoted as lstm0 in the figure) displays smooth profiles with (testing) accuracy around $88\%$. However, LSTM\_6 (denoted as LSTM6 in the figure) shows fluctuations and also does not catch up with \textit{standard} LSTM. This is an indicator that $\eta = 1\mathrm{e}{-3}$ is too large for this variant network. Since the number of parameters in LSTM6 has aggressively been reduced, it is expected that the different optimal values of $\eta$ would  work better. This is study, we consider a grid of two values around the default value. Decreasing $\eta$ to $1e-4$ improves the performance of LSTM6 to $82\%$ , however a small amount of fluctuation is still observed. Meanwhile, LSTM\_C6 did not show any improvement.

\begin{figure}[!htb]
	\centering
	\includegraphics[trim={0 0 0 0},clip,scale=0.42]{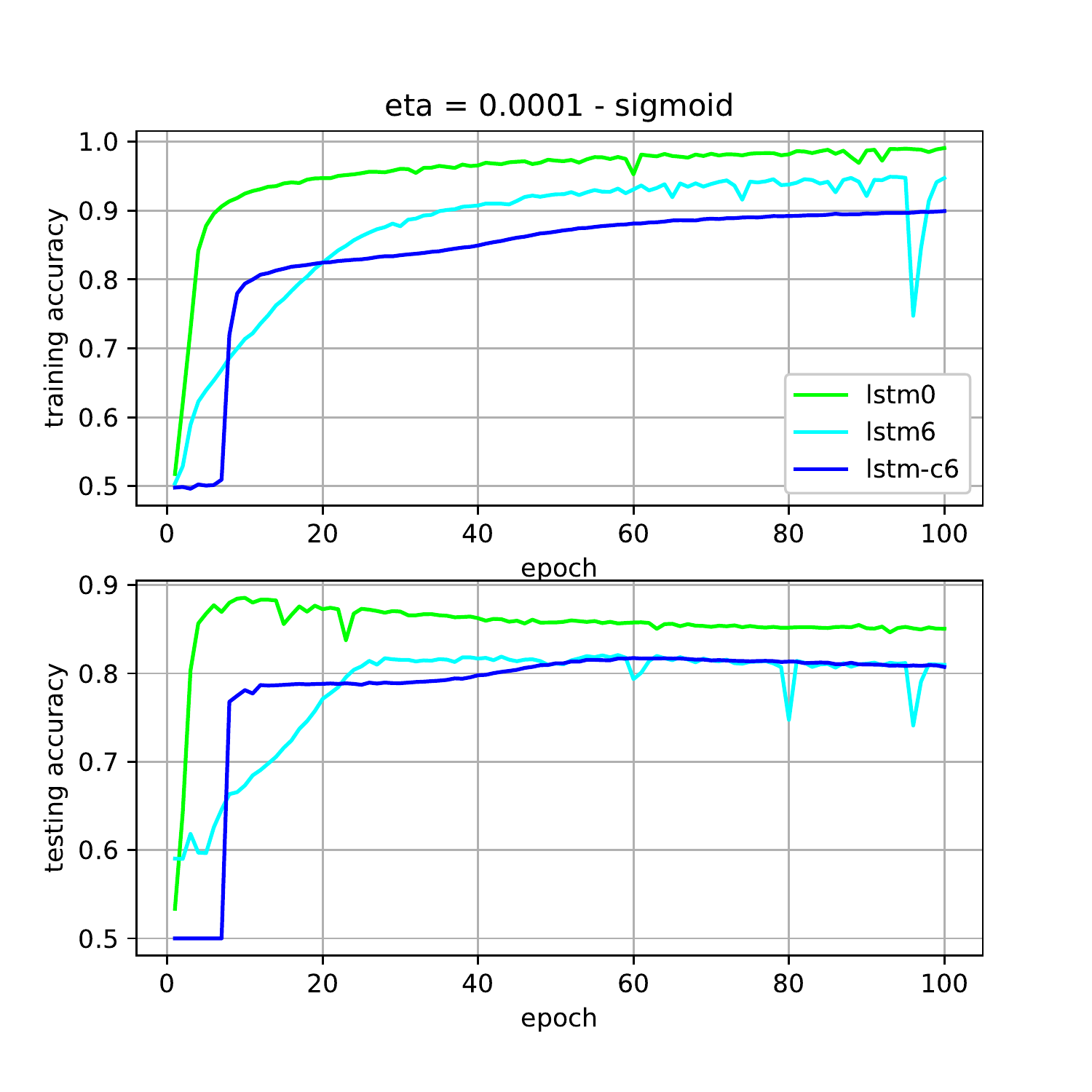}
	\caption{IMDB, Training \& Test accuracy, $\sigma=sigmoid , \eta=1\mathrm{e}{-4}$}
	\label{im0001}
\end{figure}

The typical results obtained over the eta-grid among all the epochs are shown in Table \ref{cmp-im}. 

\begin{table}[!htb]
	\caption{Best results obtained for IMDB using sigmoid}
	\centering
	\begin{tabular}{ |c|c|c|c|c| }
		\cline{3-5}
		\multicolumn{1}{c}{}
		& & $\eta = 1\mathrm{e}{-4}$ & $\eta =1\mathrm{e}{-3}$ & $\eta = 2\mathrm{e}{-3}$ \\
		\hline
		\multirow{2}{*}{LSTM} & train &  0.9906 &  1.0000 &  0.9600 \\ 
		& test &  0.8856 &  \textbf{0.8868} &  0.7775 \\ 
		\hline
		\multirow{2}{*}{LSTM6} & train &  0.9489 &  0.9387 &  0.7850 \\ 
		& test &  \textbf{0.8208} &  0.8026 &  0.7100 \\ 
		\hline
		\multirow{2}{*}{LSTM\_C6} & train &  0.8992 &  0.9445 &  0.9556 \\ 
		& test &  0.8174 &  \textbf{0.8192} &  0.7842 \\ 
		\hline
	\end{tabular}
	\label{cmp-im}
\end{table}

\subsubsection{Increasing the dimension of the state or hidden units}
To compensate for decreased number of parameters, the dimension of hidden units has been increased along with different smaller values of $\eta$. As it is shown, higher dimensions need less epoch to reach leveling off profiles. Setting $\eta = 1.25e-5$,  creates an almost fluctuation free profile. In the following figures lstm62 stands for LSTM6 using 200 hidden units.

\begin{figure}[!htb]
	\centering
	\includegraphics[trim={0 0 0 0},clip,scale=0.42]{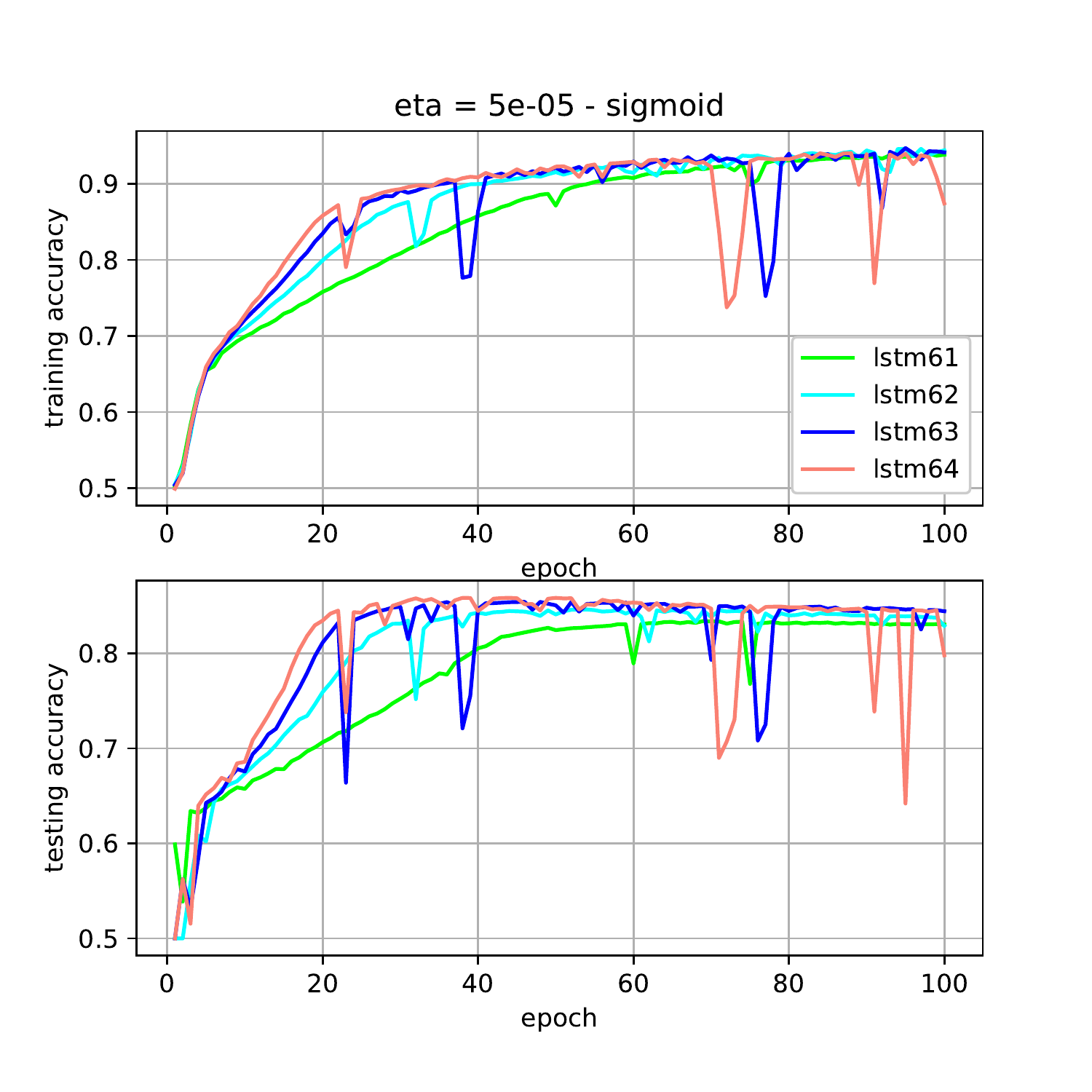}
	\caption{IMDB, Training \& Test accuracy, $\sigma=sigmoid , \eta=5\mathrm{e}{-5}$}
	\label{cmp1}
\end{figure}

\begin{figure}[!htb]
	\centering
	\includegraphics[trim={0 0 0 0},clip,scale=0.42]{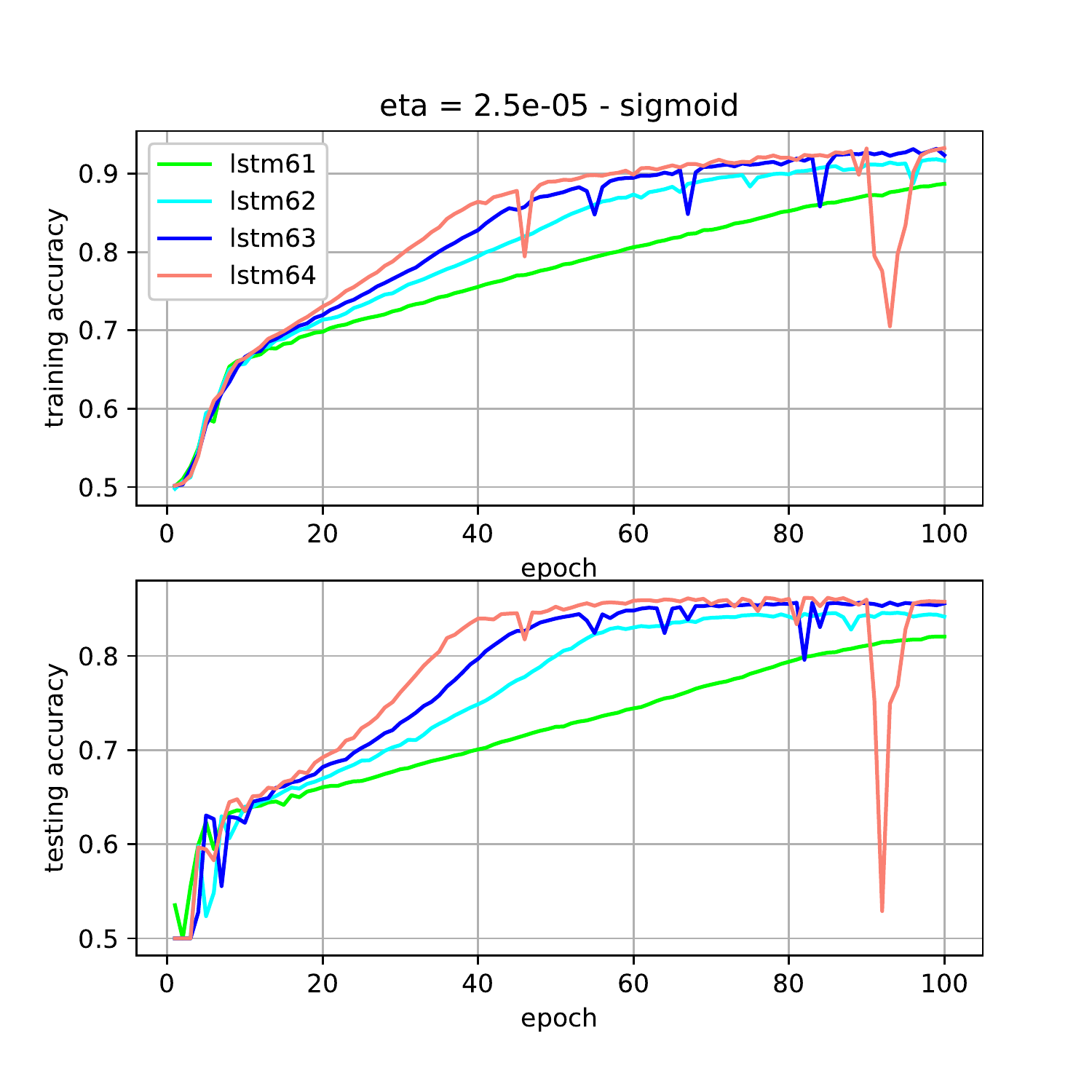}
	\caption{IMDB, Training \& Test accuracy, $\sigma=sigmoid , \eta=2.5\mathrm{e}{-5}$}
	\label{cmp2}
\end{figure}

\begin{figure}[!htb]
	\centering
	\includegraphics[trim={0 0 0 0},clip,scale=0.42]{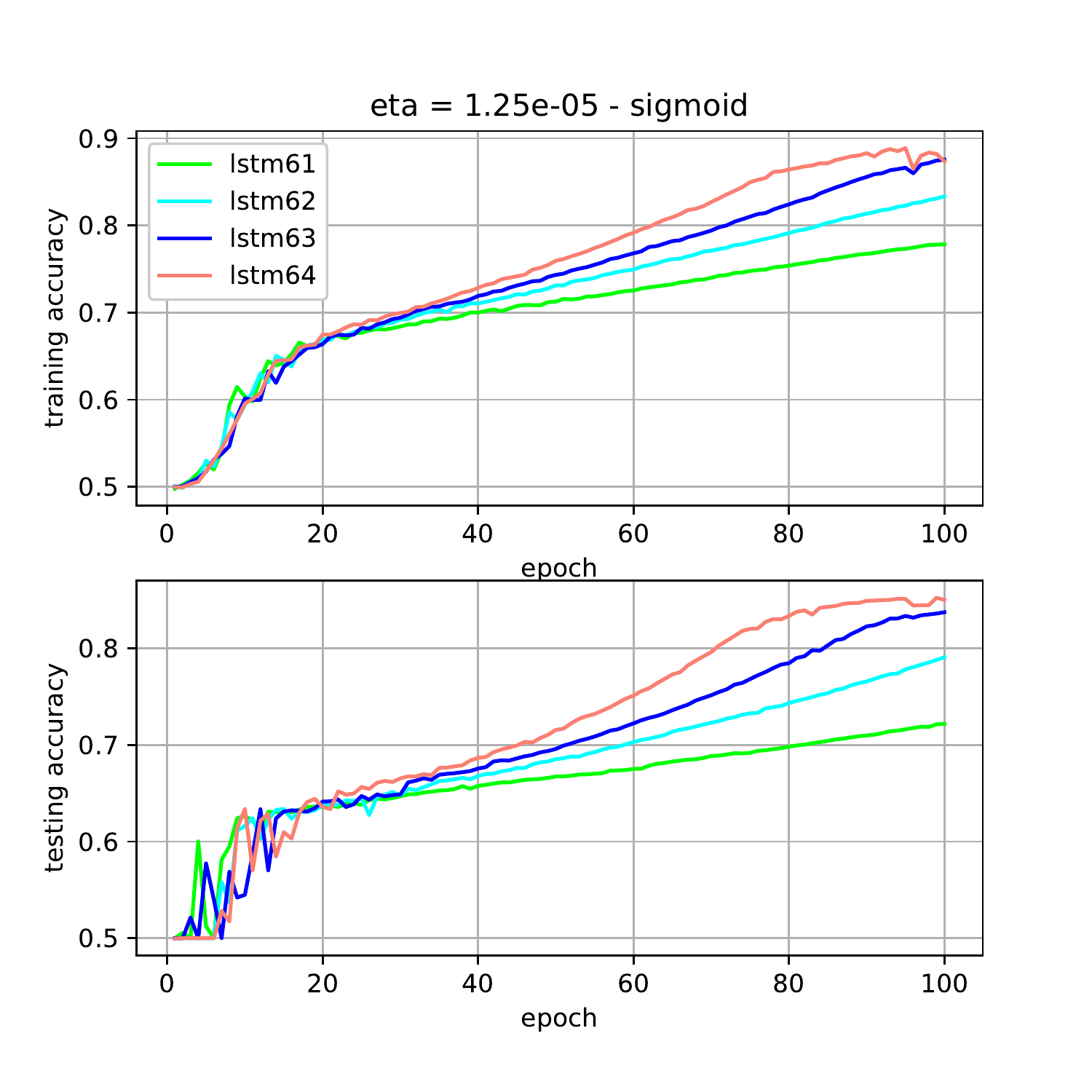}
	\caption{IMDB, Training \& Test accuracy, $\sigma=sigmoid , \eta=1.25\mathrm{e}{-5}$}
	\label{cmp3}
\end{figure}

\begin{table}[!htb]
	\caption{Best results obtained Using sigmoid.}
	\centering
	\begin{tabular}{ |c|c|c|c|c| }
		\cline{3-5}
		\multicolumn{1}{c}{}
		& & $\eta = 5\mathrm{e}{-5}$ & $\eta =2.5\mathrm{e}{-5}$ & $\eta = 1.25\mathrm{e}{-5}$ \\
		\hline
		\multirow{2}{*}{100} & train &  0.9396 &  0.8871 &  0.7783\\ 
		& test &  \textbf{0.8340} &  0.8206 &  0.7219 \\ 
		\hline
		\multirow{2}{*}{200} & train &  0.9461 &  0.9185 &  0.833 \\ 
		& test &  \textbf{0.8461} &  0.8459 &  0.7908\\ 
		\hline
		\multirow{2}{*}{300} & train &  0.9471 &  0.9319 &  0.8754\\ 
		& test &  0.8542 &  \textbf{0.8567} &  0.8374\\ 
		\hline
		\multirow{2}{*}{400} & train &  0.9404 &  0.933 &  0.8887 \\ 
		& test &  0.8585 &  \textbf{0.8618} &  0.8523\\ 
		\hline
	\end{tabular}
	\label{tanh}
\end{table}

\subsubsection{Tuning the constant forget hyper-paramter}
The forget (gate) constant value must be less than one in absolute value for bounded-input-bounded-output (BIBO) stability \cite{salem2016basic}. In our previous work \cite{AkandehSalem2017III},  $f~>~0.59$ did not work for the MNIST dataset and training would not converge. In this paper on this different dataset, we initially start with the same value (i.e., $f~=~0.59$ . To fill in the gap between \textit{standard} LSTM and LSTM6, we gradually increase the forget hyperparamter $f$ and observe that IMDB dataset produce BIBO stable performance up to $f_t=0.96$. Since the accurcy plot profiles show increasing performance trend and do not appear to level off after $100$ epochs. We run the training for $200$ epochs. It is observed that LSTM6 surpass \textit{standard} LSTM at around epoch $150$. 

\begin{figure}[!htb]
	\centering
	\includegraphics[trim={0 0 0 0},clip,scale=0.42]{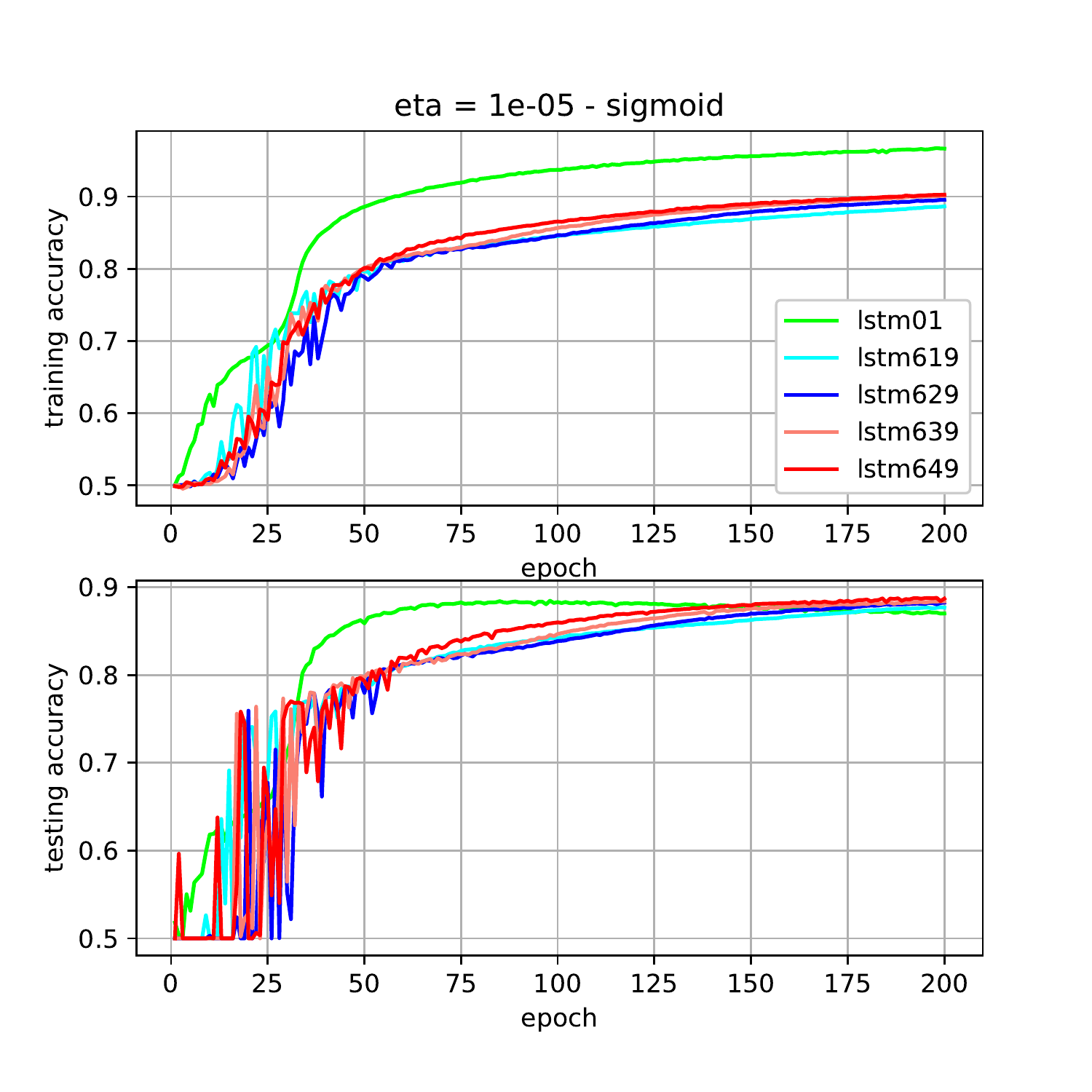}
	\caption{IMDB, Training \& Test accuracy, $\sigma=sigmoid , \eta=1\mathrm{e}{-5}$}
	\label{all9}
\end{figure}

\begin{figure}[!htb]
	\centering
	\includegraphics[trim={0 0 0 0},clip,scale=0.42]{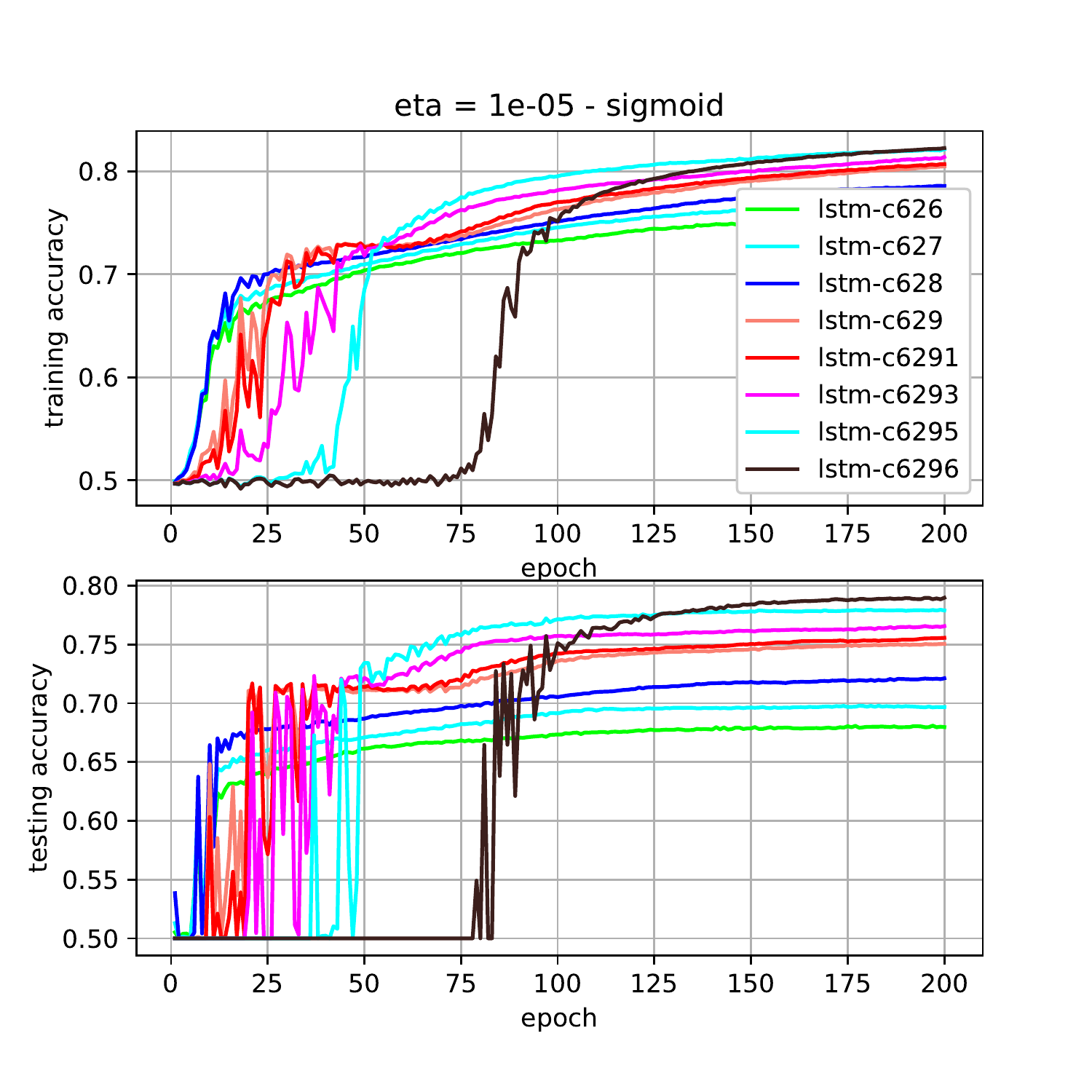}
	\caption{IMDB, Training \& Test accuracy, $\sigma=sigmoid , \eta=1\mathrm{e}{-5}$}
	\label{l6c-ft}
\end{figure}

The effect of increasing the hyper-parameter $f$ in the LSTM\_C6 network using $h=200$ and $\eta=\mathrm{e}{-5}$ is also depicted in figure \ref{l6c-ft}. In this figure, lstm-c6295 denotes LSTM\_C6 using $h=200$ and $f=0.95$.

\subsection{The20 Newsgroups dataset}
\begin{figure}[!htb]
	\centering
	\includegraphics[trim={0 0 0 0},clip, scale=0.48]{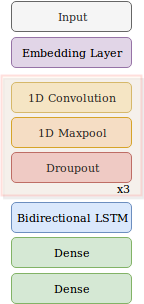}
	\vspace*{+2mm}
	\caption{News20,  Network Architecture Diagram}
	\label{n20-mdl}
\end{figure}

The 20 Newsgroups dataset is a collection of 20000 documents, containing 20 different newsgroups. GloVe embedding is used to pre-train the model \cite{keras_rowwise}. The network architecture is adapted from Keras1.2 examples. Table \ref{n20} provides the network specification. We have applied our variants in the bidirectional layer. A schematic representation of the architecture used is given in figure \ref{n20-mdl}.

\begin{table}[ht]
	\caption{Network specifications.}
	\centering
	\begin{tabular}{| c | c |} 
		\hline
		Input dimension & $1000$ \\ 
		\hline
		Embedding layer & $1000 \times 100 $ \\
		\hline
		Conv1D(128, 5,'relu') & $996 \times 128 $ \\
		\hline
		Maxpooling1D(5) & $199 \times 128 $ \\
		\hline
		Conv1D(128, 5,'relu') & $195 \times 128 $ \\
		\hline
		Maxpooling1D(5) & $39 \times 128 $ \\
		\hline
		Conv1D(128, 5,'relu') & $35 \times 128 $ \\
		\hline
		Maxpooling1D(2) & $17 \times 128 $ \\
		\hline
		Number of epochs & $ 100 $ \\
		\hline
		Bidirectional(lstmi) & 256 \\
		\hline
		Dense & 128 \\
		\hline
		Dense & 6 \\
		\hline
		Optimizer  & rmsprop  \\
		\hline
		Batch size & $128$ \\
		\hline
		Loss function & categorical cross-entropy \\
		\hline
	\end{tabular}
	\label{n20}
\end{table}

\subsubsection{The tanh activation}
Using $tanh$ as nonlinearity, the LSTM\_C6 layer results in better performance than using the LSTM6 layer and even using the \textit{standard} LSTM layer. It is observed that setting $\eta = 1\mathrm{e}{-3}$ results in test score of $79.42\%$ in LSTM\_C6 which surpasses the test score of the \textit{standard} LSTM, $77.75\%$, using $\eta = 2\mathrm{e}{-3}$. The best results obtained for the three grid values of $eta$ over100 epochs are summarized in Table \ref{tanh}.
\begin{table}[!htb]
	\caption{Best results obtained for news20 LSTM models}
	\centering
	\begin{tabular}{ |c|c|c|c|c| }
		\cline{3-5}
		\multicolumn{1}{c}{}
		& & $\eta = 5\mathrm{e}{-4}$ & $\eta =1\mathrm{e}{-3}$ & $\eta = 2\mathrm{e}{-3}$ \\
		\hline
		\multirow{2}{*}{LSTM} & train &  0.9519 &  9581 &  0.9600 \\ 
		& test &  0.7592 &  0.7750 &  \textbf{0.7775} \\ 
		\hline
		\multirow{2}{*}{LSTM6} & train &  0.8169 &  0.8448 &  0.7850 \\ 
		& test &  0.7158 & \textbf{ 0.7200} &  0.7100 \\ 
		\hline
		\multirow{2}{*}{LSTM\_C6} & train &  0.9552 &  0.9583 &  0.9556 \\ 
		& test &  0.7792 &  \textbf{0.7942} &  0.7842 \\ 
		\hline
	\end{tabular}
	\label{tanh}
\end{table}

\begin{figure}[!htb]
	\centering
	\includegraphics[trim={0 0 0 0},clip, scale=0.42]{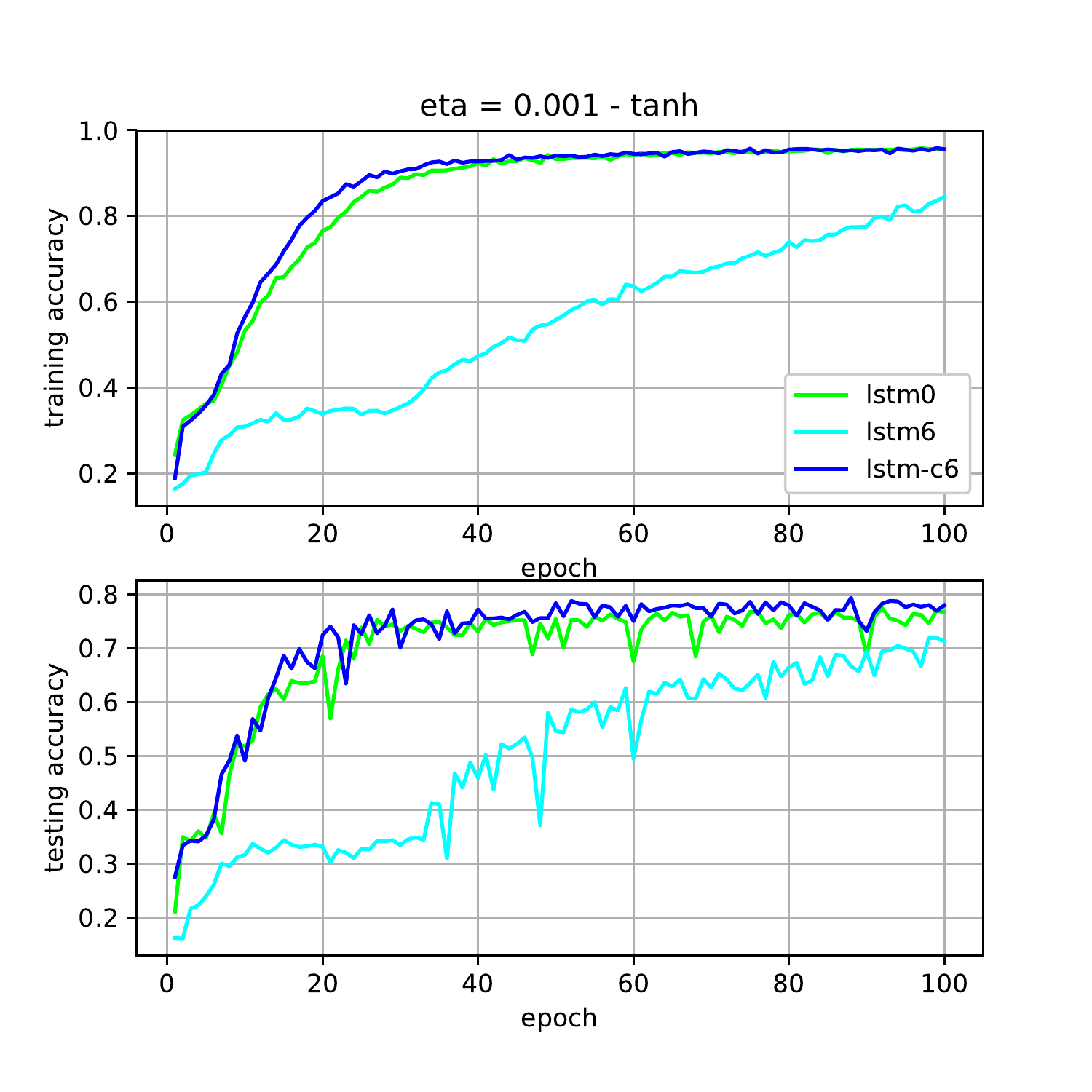}
	\caption{News20, Training \& Test accuracy, $\sigma=tanh , \eta=1\mathrm{e}{-3}$}
	\label{n-tan001}
\end{figure}

\begin{figure}[!htb]
	\centering
	\includegraphics[trim={0 0 0 0},clip, scale=0.42]{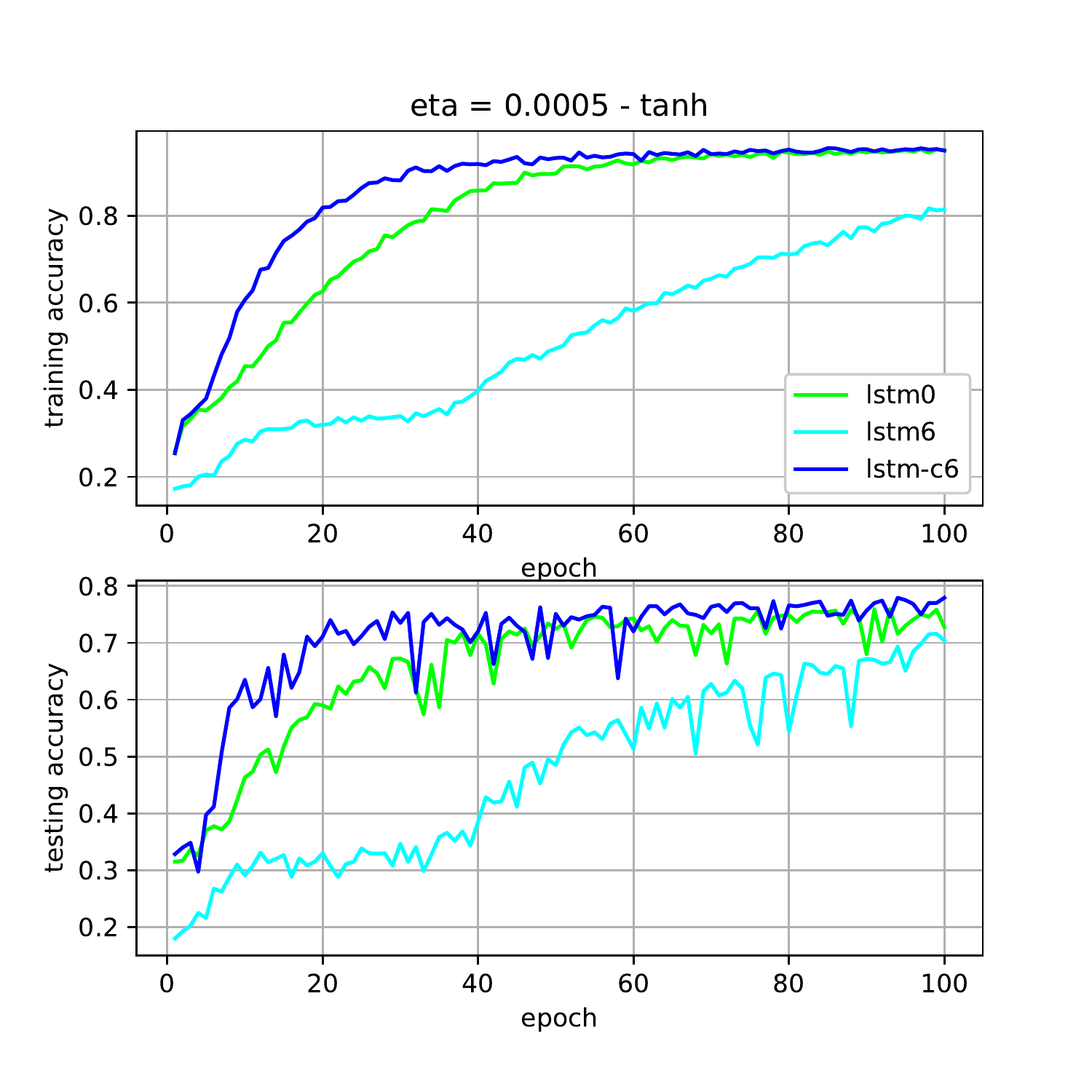}
	\caption{News20, Training \& Test accuracy, $\sigma=tanh , \eta=5\mathrm{e}{-4}$}
	\label{n-tan0005}
\end{figure}

\subsubsection{The (logistic) sigmoid activation}

Using $sigmoid$ as nonlinearity, the similar trend is observed; LSTM\_C6 shows better performance than LSTM6 and even \textit{standard} LSTM.

\begin{figure}[!htb]
	\centering
	\includegraphics[trim={0 0 0 0},clip, scale=0.42]{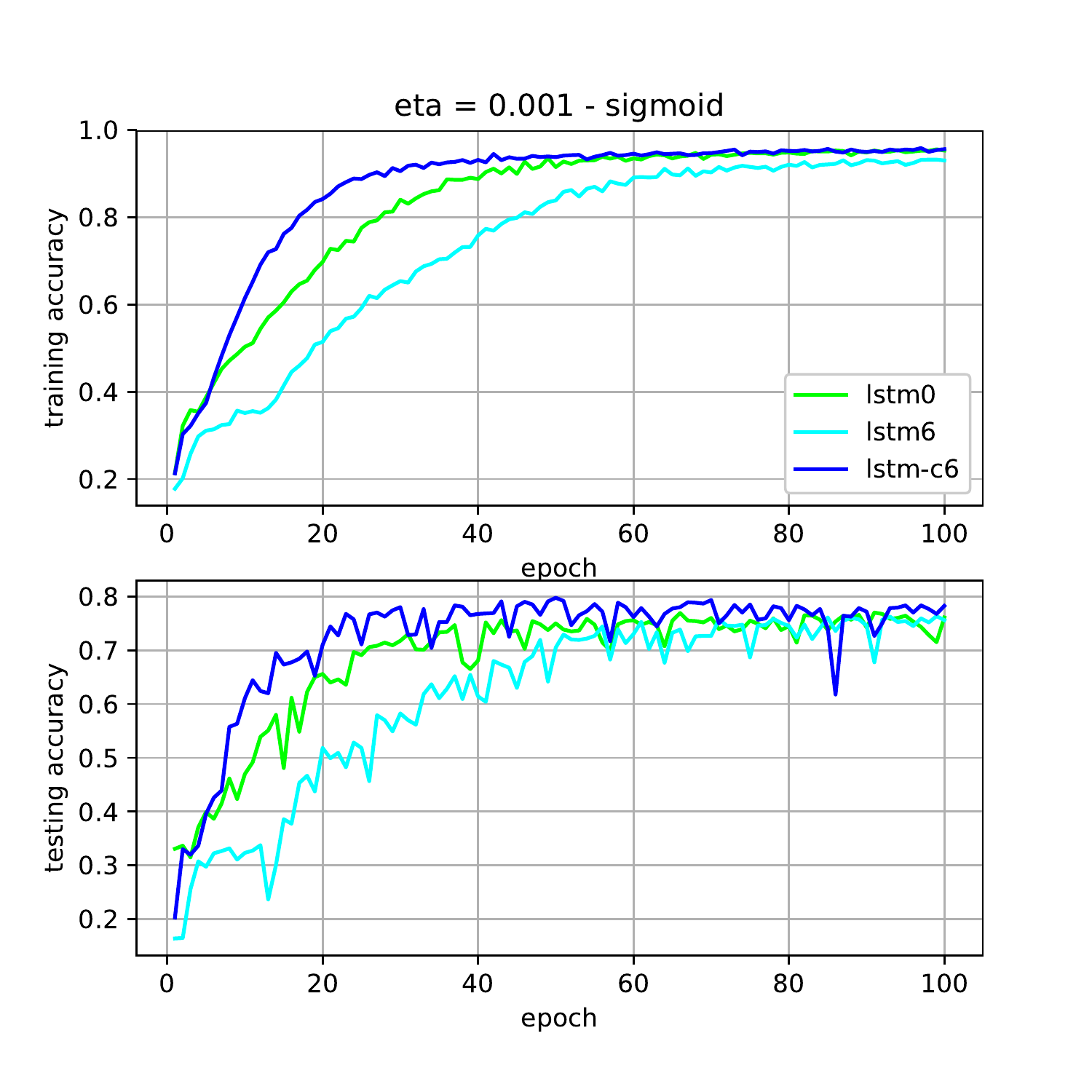}
	\caption{Training \& Test accuracy, $\sigma=sigmoid , \eta=1\mathrm{e}{-3}$}
	\label{n-sig001}
\end{figure}

\section{Conclusion}

LSTM\_6 and LSTM\_C6, which are aggressively reduced variant of the baseline \textit{standard} LSTM have been evaluated on the benchmark classical IMDB and 20 Newsgroups datasets. In these slim LSTM variants, the gates are set at constants, and effectively only the forget gate serves now as a hyper-parameter to ensure BIBO stability of the discrete dynamic recurrent neural network (RNN).  LSTM\_C6 further reduced the matrix $U_c$ in the input block equation into a vector with point-wise (Hadamard) multiplication. We tried limited grid of 3 values of the learning rate centered around a default value for the \textit{standard} LSTM RNN using in the Keras Library. Moreover, the network dimension can be used as a hyper-parameter to improved the slim LSTM variants. These investigations have shown that the capacity of the slim LSTMS can match the \textit{standard} LSTM while still saving computational expense. It was observed that as we increase the number of hidden units the performance improves. Finally, using the hyper-parameter $f$ in place of the forget gate $f_t$, the training/ testing performance can also improve, up to to the value $f =0.96$ for the IMDB dataset. This enables LSTM6 to surpass \textit{standard} LSTM at around 150 epochs. In the 20 Newsgroups dataset, LSTM\_C6 surpasses base LSTM without much parameter tuning. As a results we conclude that these simplified models are comparable to the \textit{standard} LSTM. Thus, these slim LSTM variants may be suitably employed in applications in order to benefit from realtime speed and/or computational expense.

\section*{Acknowledgment}
This work was supported in part by the National Science Foundation under grant No. ECCS-1549517. 


\end{document}